\documentclass[letterpaper]{article}
\usepackage{aaai17}
\usepackage{times}
\usepackage{helvet}
\usepackage{courier}
\usepackage{amsmath}
\usepackage{multirow}
\usepackage{url}
\usepackage[font=normalsize]{subfig}
\usepackage[noend]{algpseudocode}

\usepackage{amssymb}
\usepackage{bm}
\usepackage{balance}
\usepackage{arydshln}
\usepackage{mathtools}

\usepackage{amssymb}
\usepackage{textcomp}

\usepackage{xspace}
\usepackage{color}

\newcommand{\eg}{\emph{e.g.,}\xspace}
\newcommand{\ie}{\emph{i.e.,}\xspace}
\newcommand{\comments}[1]{}

\DeclareMathOperator*{\argmax}{arg\,max}

\def\year{2017}\relax

\begin{document}
%
\title{Neural Machine Translation with Reconstruction}
\def\sstaff{$^\ddag$}
\def\fndaff{$^\dagger$}
\author{Zhaopeng Tu\fndaff ~~~ Yang Liu\sstaff ~~~ Lifeng Shang\fndaff ~~~ Xiaohua Liu\fndaff ~~~ Hang Li\fndaff
\\
\\
{ \fndaff {Noah's Ark Lab, Huawei Technologies, Hong Kong}}   \\
{ \tt \{tu.zhaopeng,shang.lifeng,liuxiaohua3,hangli.hl\}@huawei.com}\\
{ \sstaff {Department of Computer Science and Technology, Tsinghua University, Beijing}}\\
{ \tt liuyang2011@tsinghua.edu.cn}\\
}
\maketitle

\begin{abstract}
Although end-to-end Neural Machine Translation (NMT) has achieved remarkable progress in the past two years, it suffers from a major drawback: translations generated by NMT systems often lack of adequacy. It has been widely observed that NMT tends to repeatedly translate some source words while mistakenly ignoring other words.  To alleviate this problem, we propose a novel {\em encoder-decoder-reconstructor} framework for NMT. 
The reconstructor, incorporated into the NMT model, manages to reconstruct the input source sentence from the hidden layer of the output target sentence, to ensure that the information in the source side is transformed to the target side as much as possible.
Experiments show that the proposed framework significantly improves the adequacy of NMT output and achieves superior translation result over state-of-the-art NMT and statistical MT systems.
\end{abstract}

\section{Introduction}

Past several years have observed a significant progress in Neural Machine Translation (NMT)~\cite{Kalchbrenner:2013:EMNLP,Cho:2014:EMNLP,Sutskever:2014:NIPS,Bahdanau:2015:ICLR}.
Particularly, NMT has significantly enhanced the performance of translation between a language pair involving rich morphology prediction and/or significant word reordering~\cite{Luong:2015:IWSLT,Bentivogli:2016:EMNLP}. Long Short-Term Memory \cite{Hochreite:1997} enables NMT to conduct long-distance reordering, which is a significant challenge for Statistical Machine Translation (SMT)~\cite{Brown:1993:CL,Koehn:2003:NAACL}.

Unlike SMT which employs a number of components, NMT adopts an end-to-end {\em encoder-decoder} framework to model the entire translation process. The role of encoder is to summarize the source sentence into a sequence of latent vectors, and the decoder acts as a language model to generate a target sentence word by word by selectively leveraging the information from the latent vectors at each step. 
In learning, NMT essentially estimates the likelihood of a target sentence given a source sentence.

However, conventional NMT faces two main problems:
\begin{itemize}

\item[1] {\em Translations generated by NMT systems often lack of adequacy}.
When generating target words, the decoder often repeatedly selects some parts of the source sentence while ignoring other parts, which leads to over-translation and under-translation~\cite{Tu:2016:ACL}. This is mainly due to that NMT does not have a mechanism to ensure that the information in the source side is completely transformed to the target side.
\item[2] {\em Likelihood objective is suboptimal in decoding.} NMT utilizes a beam search to find a translation that maximizes the likelihood. However, we observe that likelihood favors short translations, and thus fails to distinguish good translation candidates from bad ones in a large decoding space (\eg beam size $= 100$). The main reason is that likelihood only captures unidirectional dependency from source to target, which does not correlate well with translation adequacy~\cite{Li:2016:NAACL,Shen:2016:ACL}.
\end{itemize}

\begin{figure}[t]
\centering
\includegraphics[width=0.4\textwidth]{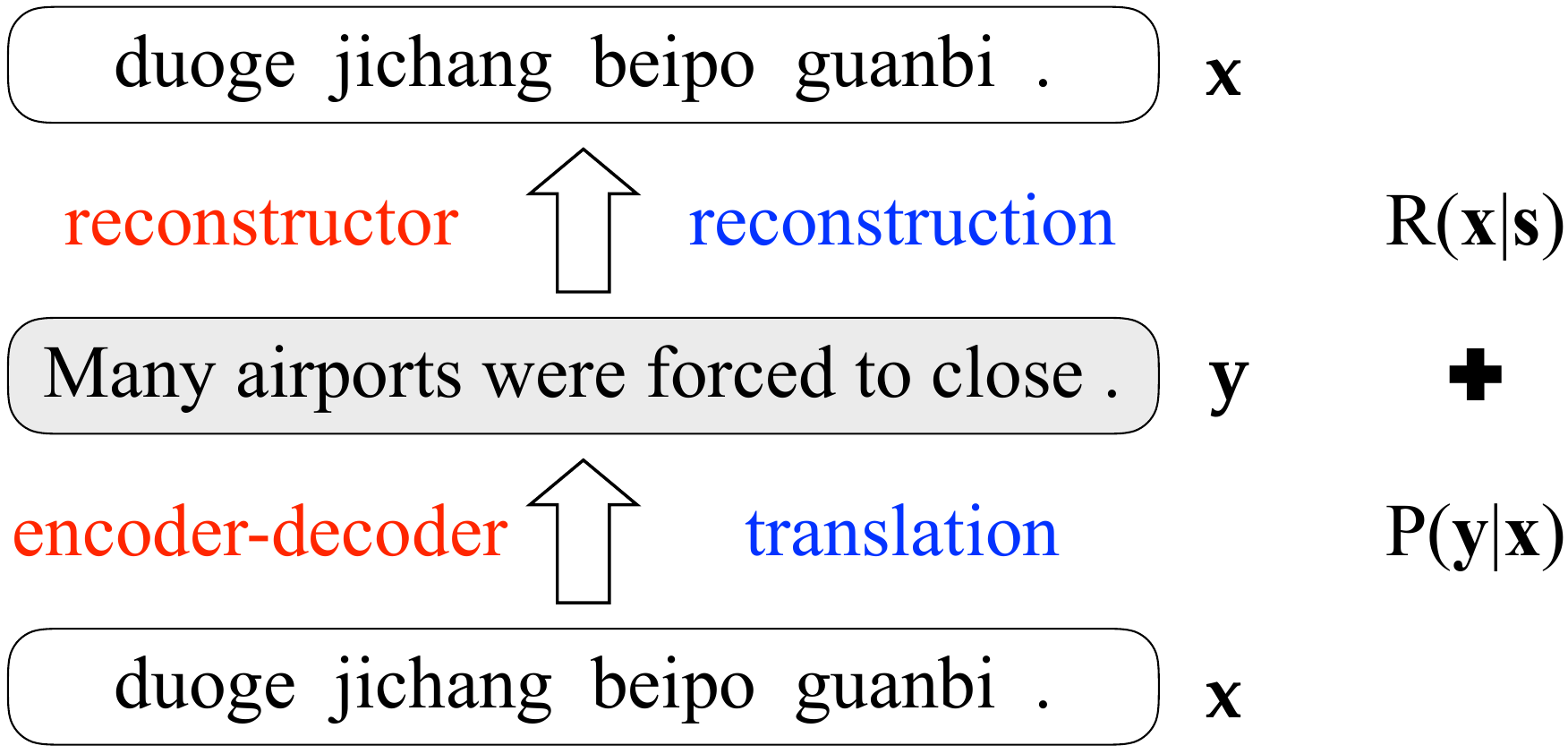}
\caption{Example of NMT with reconstruction. Our idea is to leverage reconstruction score $R({\bf x}|{\bf s})$ as an auxiliary objective to measure the adequacy of translation candidate, where ${\bf s}$ is the target-side hidden layer in decoder for generating the translation ${\bf y}$. Linear interpolation of likelihood score $P({\bf y}|{\bf x})$ and reconstruction score is used to (1) improve parameter learning for generating better translation candidates in {\em training}, and (2) conduct better rerank of generated candidates in {\em testing}.}
\label{figure-example}
\end{figure}

While previous work partially solves the above problems, in this work we propose a novel {\em encoder-decoder-reconstructor} model for NMT, aiming at alleviating these problems in a unified framework. 
As shown in Figure~\ref{figure-example}, given a Chinese sentence ``duoge jichang beipo guanbi .'', the standard encoder-decoder translates it into an English sentence and assigns a likelihood score. Then, the newly added reconstructor reconstructs the translation back to the source sentence and calculates the corresponding reconstruction score. Linear interpolation of the two scores produces an overall score of the translation.

As seen, the added reconstructor imposes a constraint that an NMT model should be able to reconstruct the input source sentence from the target-side hidden layers, which encourages decoder to embed complete information of the source side.
The reconstruction score serves as an auxiliary objective to measure the adequacy of translation.
The combined objective consisting of likelihood and reconstruction, which measures both fluency and adequacy of translations, is used in both training and testing.

Experimental results show that the proposed approach consistently improves the translation performance when increasing the decoding space. Our model achieves a significant improvement of 2.3 BLEU points over a strong attention-based NMT system, and of 4.5 BLEU points over a state-of-the-art SMT system, trained on the same data.

\section{Background}
\label{sec-background}

\subsection{Encoder-Decoder based NMT}
\label{sec-nmt}

Given a source sentence ${\bf x}=x_1, \dots x_j, \dots x_J$ and a target sentence ${\bf y}=y_1, \dots y_i, \dots y_{I}$, end-to-end NMT directly models the translation probability word by word:
\begin{equation}
P({\bf y}|{\bf x}) = \prod_{i=1}^{I} P(y_i| y_{<i}, {\bf x}; \theta)
\end{equation}
where $\theta$ is the model parameters and $y_{<i}=y_1,\dots, y_{i-1}$ is partial translation.
Prediction of the {\em i}-th target word is generally made in an {\em encoder-decoder} framework:
\begin{equation}
P(y_i| y_{<i}, {\bf x}; \theta) \propto \exp\Big\{ f(y_{i-1}, s_i, c_i; \theta) \Big\}
\end{equation}
where $s_i$ is the $i$-th hidden target state computed by the decoder Recurrent Neural Network (RNN), $c_i$ is the $i$-th source representation for generating the $i$-th target word, and $f(\cdot)$ is an activation function in the decoder. Current NMT models differ in their ways of calculating $c_i$ from the hidden states from the encoder. Please refer to~\cite{Sutskever:2014:NIPS,Bahdanau:2015:ICLR} for more details.
The parameters of NMT model are trained to maximize the {\em likelihood} of a set of training examples $\{\left[{\bf x}^n, {\bf y}^n\right]\}_{n=1}^{N}$:
\begin{equation}
\mathcal{L}(\theta) = \argmax_{\theta}\sum_{n=1}^{N}\log P({\bf y}^n|{\bf x}^n; \theta)
\label{eqn-standard-training}
\end{equation}

When generating each target word, the decoder adaptively selects partial information (\ie $c_i$) from the encoder. This actually adopts a {\em greedy way} to select the most useful information for each generated word. There is, however, no mechanism to guarantee that the decoder conveys complete information from the source sentence to the target sentence.

\begin{table}[t]
\centering
\renewcommand\arraystretch{1.1}
\begin{tabular}{l||c|c|c|c}
    \multirow{2}{*}{\bf Beam}	&	\multicolumn{2}{c|}{\bf Likelihood}	&	\multicolumn{2}{c}{\bf + Normalization}\\
    \cline{2-5}
    	&	BLEU	&	Length	&	BLEU	&	Length\\
    \hline
    10		&	35.46	&	29.9	&	34.51	&	32.3\\
    100		&	25.80	&	17.9	&	33.39	&	32.7\\
    1000	&	1.38	&	4.6	&	29.50	&	33.9\\
\end{tabular}
\caption{Likelihood favors short translation candidates (``{\em Length}''), and thus cannot further improve translation performance (``{\em BLEU}'') as the decoding space (``{\em Beam}'') increases. Normalizing likelihood by candidate length (``{\em Normalization}'') does not solve the problem.} 
\label{table-length}
\end{table}

In addition, we find that the performance of NMT decreases as the decoding space increases, as shown in Table~\ref{table-length}.
This is because likelihood favors short but inadequate translation candidates, which are newly added together with good candidates in larger decoding spaces.
Normalizing likelihood by translation length faces the same problem.

It is important to  introduce an auxiliary objective to measure the adequacy of translation, which complements likelihood.

\subsection{Reconstruction in Auto-Encoder}

Reconstruction is a standard concept in auto-encoder, which is usually realized by a feed forward network~\cite{Bourlard:1988:BC,Vincent:2010:JLMR,Socher:2011:EMNLP}. 
The model consists of an encoding function to compute a representation from an input, and a decoding function to reconstruct the input from the representation.
The parameters involved in the two functions are trained to maximize the {\em reconstruction score}, which measures the similarity between the original input and reconstructed input.

Reconstruction examines whether the reconstructed input is faithful to the original input, which is essentially similar to the consideration of adequacy in translation. It is natural to integrate reconstruction into NMT to enhance adequacy of translation.
The basic idea of our approach is to reconstruct the source sentence from the latent representations of decoder, and use the reconstruction score as the adequacy measure.
Analogous to auto-encoder, our approach also learns a latent representation of source sentence on the target side. 
Our approach can be viewed as a {\em supervised auto-encoder} in the sense that the latent representation is not only used to reconstruct the source sentence, but also used to generate the target sentence.

\section{Approach}

\subsection{Architecture}

\begin{figure}[t]
\centering
\includegraphics[width=0.3\textwidth]{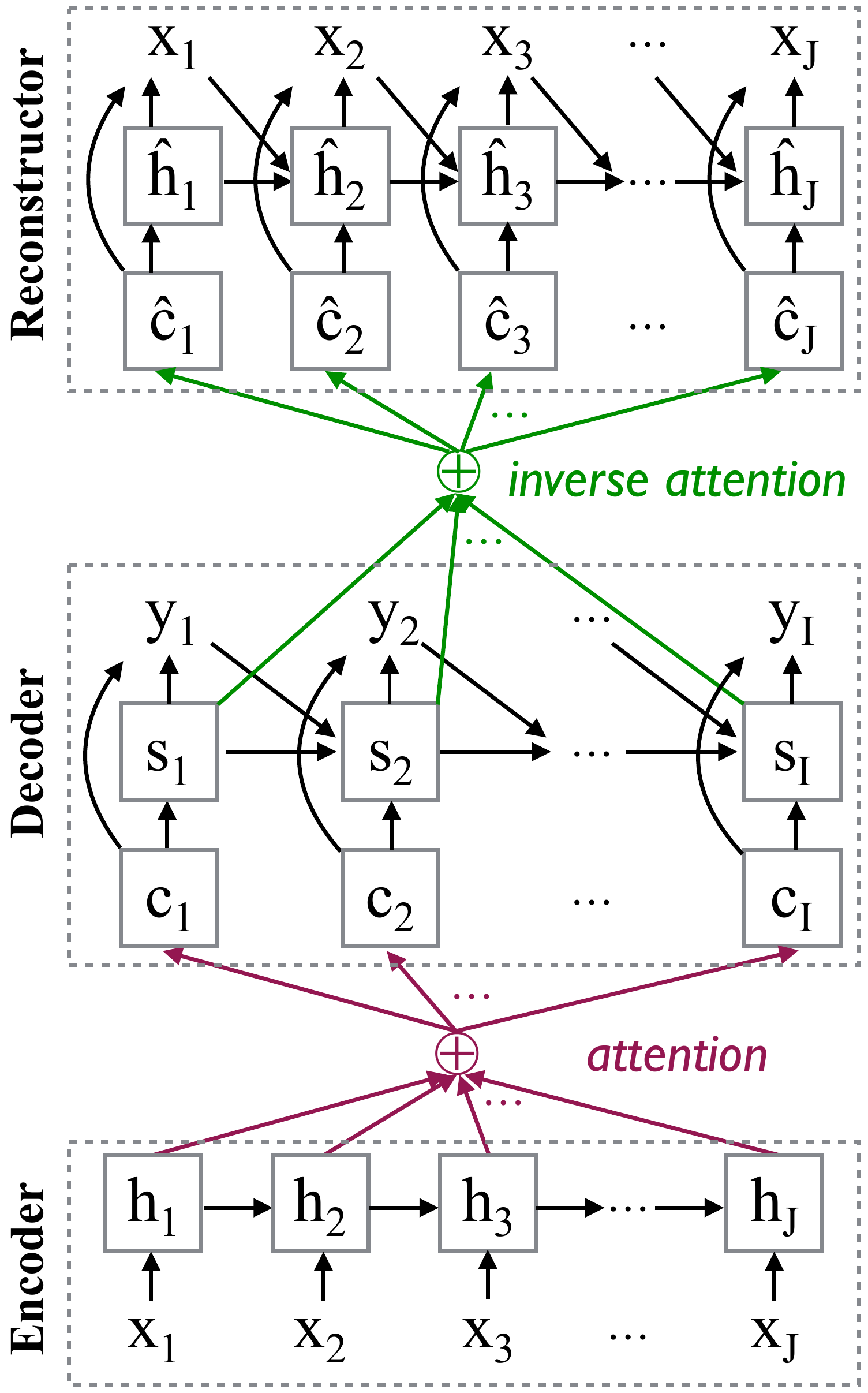}
\caption{Architecture of NMT with reconstruction, which introduces a reconstructor to map from the hidden layer at the target side to the original input.}
\label{figure-architecture}
\end{figure}

We prepose a novel {\em encoder-decoder-reconstructor} framework. More specifically, we base our approach on top of attention-based NMT~\cite{Bahdanau:2015:ICLR,Luong:2015:EMNLP}, which will be used as baseline in the experiments later.
We note that the proposed approach is generally applicable to any other type of NMT architectures, such as the sequence-to-sequence model~\cite{Sutskever:2014:NIPS}.
The model architecture, shown in Figure~\ref{figure-architecture}, consists of two components:
\begin{itemize}
\item Standard {\em encoder-decoder} reads the input sentence and outputs its translation along with the likelihood score, as shown in the background section.
\item Added {\em reconstructor} reads the hidden state sequence from the decoder and outputs a score of exactly reconstructing the input sentence, which we will describe below.
\end{itemize}

\paragraph{Reconstructor}
As shown in Figure~\ref{figure-architecture}, the reconstructor reconstructs the input. Here we use the hidden layer at the target side as the representation of the translation, since it plays a key role in generation of the translation. We aim at encouraging it to embed complete source information, and in the meantime to reduce the complexity of model and make the training easy.

Specifically, the reconstructor reconstructs the source sentence word by word, which is conditioned on the inverse context vector $\hat{c}_j$ for each input word $x_j$. The inverse context vector $\hat{c}_j$ is computed as a weighted sum of hidden layers ${\bf s}$ at the target-side:
\begin{equation}
\hat{c}_j = \sum_{i=1}^{I}{\hat{\alpha}_{j,i}\cdot s_i}
\end{equation}
The weight $\hat{\alpha}_{j,i}$ of each hidden layer $s_j$ is computed by an added inverse attention model, which has its own parameters independent from the original attention model.
The reconstruction probability is calculated by
\begin{eqnarray}
R({\bf x}|{\bf s}) &=& \prod_{j=1}^{J} R(x_j| x_{<j}, {\bf s}) \nonumber \\
                                &=& \prod_{j=1}^{J} g_r(x_{j-1}, \hat{h}_j, \hat{c}_j) 
\end{eqnarray}
where $\hat{h}_j$ is the hidden state in the reconstructor, and computed by
\begin{eqnarray}
\hat{h}_j &=& f_r(x_{j-1}, \hat{h}_{j-1}, \hat{c}_j)
\label{eqn-hidden-state}
\end{eqnarray}
Here $g_r(\cdot)$ and $f_r(\cdot)$ are softmax function and activation function for the reconstructor, respectively. The source words ${\bf x}$ share the same word embeddings with the encoder.

\subsection{Training}
Formally, we train both the encoder-decoder $P({\bf y}|{\bf x}; \theta)$ and the reconstructor $R({\bf x} | {\bf s}; \gamma)$ on a set of training examples $\{\left[{\bf x}^n, {\bf y}^n\right]\}_{n=1}^{N}$, where ${\bf s}$ is the state sequence in the decoder after generating ${\bf y}$, and $\theta$ and $\gamma$ are model parameters in the encoder-decoder and reconstructor respectively. The new training objective is:
\begin{eqnarray}
J(\theta, \gamma) = \argmax_{\theta, \gamma}  \sum_{n=1}^{N} \bigg\{  \underbrace{\log P({\bf y}^n|{\bf x}^n; \theta)}_\text{\normalsize \em likelihood}  \nonumber \\  
                                                                                                                                   +  \lambda \underbrace{\log R({\bf x}^n | {\bf s}^n; \gamma)}_\text{\normalsize \em reconstruction} \bigg\}
\end{eqnarray}
where $\lambda$ is a hyper-parameter that balances the preference between likelihood and reconstruction.

Note that the objective consists of two parts: likelihood measures translation fluency, and reconstruction measures translation adequacy.
It is clear that the combined objective is more consistent with the goal of enhancing overall translation quality, and can more effectively guide the parameter training for making better translation.

\subsection{Testing}

\begin{figure}[t]
\centering
\includegraphics[width=0.35\textwidth]{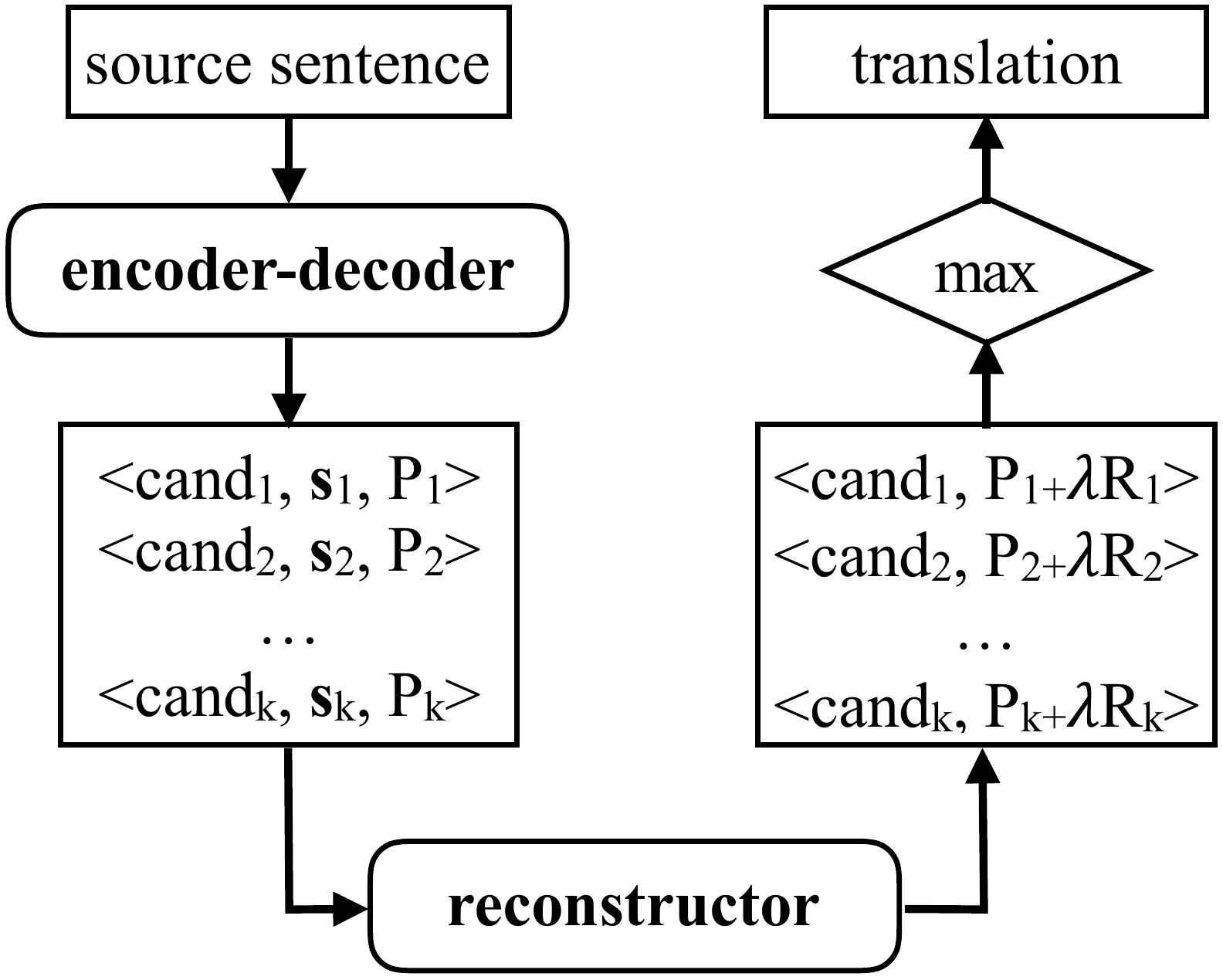}
\caption{Illustration of testing with reconstructor.}
\label{figure-testing}
\end{figure}

Once a model is trained, we use a beam search to find a translation that approximately maximizes both the likelihood and reconstruction score. As shown in Figure~\ref{figure-testing}, given an input sentence, a two-phase scheme is used:
\begin{itemize}
  \item[1] The standard encoder-decoder produces a set of translation candidates, each of which is a triple consisting of a translation candidate, its corresponding hidden layer at the target-side ${\bf s}$, and its likelihood score  $P$.
  \item[2] For each translation candidate, the reconstructor reads its corresponding hidden layer at the target-side and outputs an auxiliary reconstruction score $R$. Linear interpolation of likelihood $P$ and reconstruction score $R$ produces an overall score, which is used to select the final translation.\footnote{Interpolation weight $\lambda$ in testing is the same as in training.}
\end{itemize}
In testing, reconstruction works as a reranking technique to select a better translation from the $k$-best candidates generated by the decoder.

\section{Experiments}

\subsection{Setup}

We carry out experiments on Chinese-English translation.
The training dataset consists of 1.25M sentence pairs extracted from LDC corpora, with 27.9M Chinese words and 34.5M English words respectively.\footnote{The corpora include LDC2002E18, LDC2003E07, LDC2003E14, LDC2004T07, LDC2004T08 and LDC2005T06.}
We choose the NIST 2002 (MT02) dataset as validation set, and the NIST 2005 (MT05), 2006 (MT06) and 2008 (MT08) datasets as test sets.
We use the case-insensitive 4-gram NIST BLEU score~\cite{Papineni:2002} as evaluation metric, and \emph{sign-test}~\cite{Collins:2005} for statistical significance test.

We compare our method with state-of-the-art SMT and NMT models:
\begin{itemize}
    \item \textsc{Moses}
    ~\cite{Koehn:2007:ACL}: an open source phrase-based translation system with default configuration and a 4-gram language model trained on the target portion of training data.
    \item \textsc{RNNSearch}: our re-implemented attention-based NMT system, which incorporates dropout~\cite{Hinton:2012:arXiv} on the output layer and improves the attention model by feeding the lastly generated word.
\end{itemize}

For training \textsc{RNNSearch}, we limit the source and target vocabularies to the most frequent 30K words in Chinese and English.
We train each model with the sentences of length up to 80 words in the training data. 
We shuffle mini-batches as we proceed and the mini-batch size is 80.
The word embedding dimension is 620 and the hidden layer dimension is 1000.
We train for 15 epochs using Adadelta~\cite{Zeiler:2012:arXiv}. 

For our model, we use the same setting as \textsc{RNNSearch} if applicable.  We set the hyper-parameter $\lambda = 1$.  
The parameters of our model (\ie encoder and decoder, except those related to reconstructor) are initialized by the \textsc{RNNSearch} model trained on a parallel corpus. 
We further train all the parameters of our model for another 10 epochs.

\subsection{Correlation between Reconstruction and Adequacy}

\vspace{-5pt}
\begin{table}[h]
\centering
\renewcommand\arraystretch{1.1}
\begin{tabular}{l|c|c}
    			&	{\bf Adequacy}	&	{\bf Fluency}\\
    \hline
    Evaluator1	&	0.514		&	0.301\\
    Evaluator2	&	0.499		&	0.307\\
\end{tabular}
\caption{Correlation between reconstruction score and translation adequacy (and fluency).} 
\label{table-correlation}
\vspace{-5pt}
\end{table}

\noindent In the first experiment, we investigate the validity of our assumption that reconstruction score correlates well with translation adequacy, which is the underlying assumption of the approach. 
We conduct a subjective evaluation: two human evaluators are asked to evaluate the translations of 200 source sentences randomly sampled from the test sets. 
We calculate Pearson Correlation between the reconstruction scores and the corresponding adequacy and fluency scores on the samples, as shown in Table~\ref{table-correlation}. Two evaluators produce similar results: reconstruction score is more related to translation adequacy than fluency.

\subsection{Effect of Reconstruction on Translation}

\begin{figure}[h]
\begin{center}
            \includegraphics[width=0.45\textwidth]{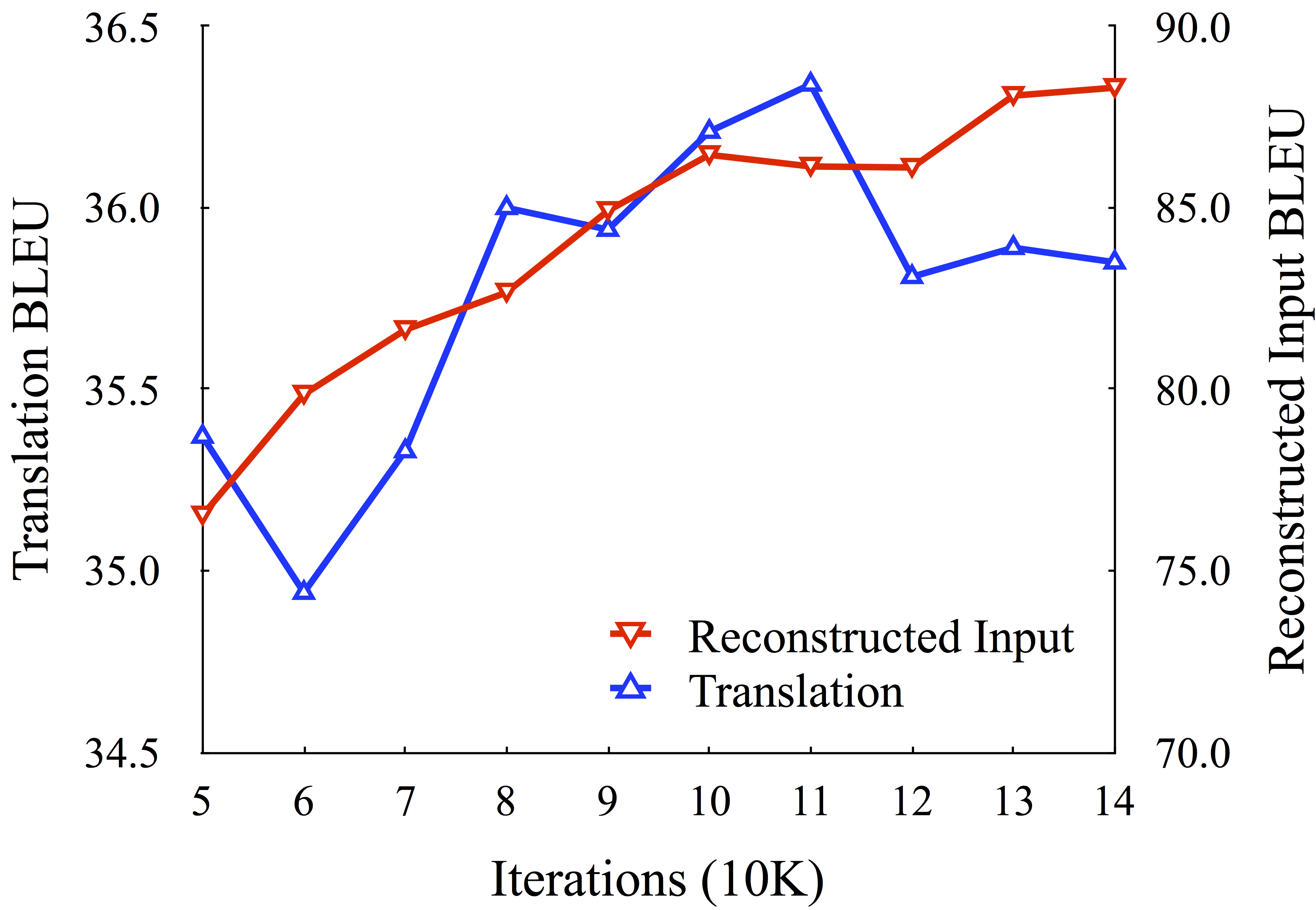}
      \caption{{\bf Learning curves} --  translation (left $y$-axis) and reconstruction (right $y$-axis) performances (in BLEU scores) on the validation set as training progresses. We find our approach is capable of generating better translations over time by better reconstructing original inputs.}
      \label{figure-reconstruction}
  \end{center}
  \vspace{-10pt}
\end{figure}

In this experiment, we investigate the effect of reconstruction on translation performance over time, which is measured in BLEU scores on the validation set.
For reconstruction, we use the reconstructor to stochastically generate a source sentence for each translation,\footnote{Note that it is different from the standard procedure, which calculates the probability of exactly reconstructing the original input.}
and calculate the BLEU score of the reconstructed input under the reference of the original input.
Generally, as shown in Figure~\ref{figure-reconstruction}, the BLEU score of translation goes up with the improvement of reconstruction over time.
The translation performance reaches a peak at iteration 110K, when the model achieves a balance between likelihood and reconstruction score. 
Therefore, we use the trained model at iteration 110K in the following experiments.

\begin{table*}[t]
\setcounter{table}{3}
\centering
\renewcommand\arraystretch{1.1}
\begin{tabular}{l|c||l|llll|l}
{\bf Model}	&	{\bf Beam}	&	{\bf Tuning}	&	{\bf MT05}  &  {\bf MT06}	  &	{\bf MT08}  &  {\bf All}	&	{\bf Oracle}\\
    \hline
   Moses	              	&	100	&	34.03&	31.37	&	30.85	&	23.01	&	28.44	&	{\em 35.17}\\
   \hline
   \hline
   \multirow{2}{*}{\textsc{RNNSearch}}		&	10	&	35.46	&	32.63	&	32.85	&	25.96	&	30.65	&	{\em 34.27}\\
    						&	100	&	33.39	&	29.58	&	30.21	&	23.76	&	27.97	&	{\em 40.20}\\
   \hline
   \multirow{2}{*}{\textsc{RNNSearch}+Reconstruction}		&	10	&	36.34$^{*}$	&	33.73$^{*}$	&	34.15$^{*}$	&	26.85$^{*}$	&	31.73$^{*}$	&	{\em 36.05}\\
    										&	100	&	37.35$^{*}$	&	34.88$^{*}$	&	35.19$^{*}$	&	27.93$^{*}$	&	32.94$^{*}$	&	{\em 42.49}\\
\end{tabular}
\caption{Evaluation of translation quality. ``Oracle'' is the best BLEU score that the $k$-best translation candidates can achieve on all the test sets.
``*'' indicate statistically significant difference ($p < 0.01$) from the baseline ``\textsc{RNNSearch} (Beam=10)''.}
\label{table-translation-results}
\end{table*}

\subsection{Effect of Reconstruction in Large Decoding Space}

\begin{table}[h]
\setcounter{table}{2}
\centering
\renewcommand\arraystretch{1.1}
\begin{tabular}{l||c|c|c|c}
    \multirow{2}{*}{\bf Beam}	&	\multicolumn{2}{c|}{\bf Likelihood}	&	\multicolumn{2}{c}{\bf +Reconstruction}\\
    \cline{2-5}
    	&	BLEU	&	Length	&	BLEU	&	Length\\
    \hline
    10		&	35.46	&	29.9	&	36.34	&	29.1\\
    100		&	25.80	&	17.9	&	37.35	&	27.1\\
    1000	&	1.38	&	4.6	&	37.77	&	26.6\\
\end{tabular}
\caption{Translation performances of different decoding beams on the validation set, in which the averaged length of reference translations is $27.0$.}
\label{table-reconstruction-length}
\vspace{-10pt}
\end{table}

Can our approach cope with the limitation of likelihood in large decoding spaces? To answer this question, we investigate the effect of reconstruction on different beam sizes $k$, as shown in Table~\ref{table-reconstruction-length}.
Our approach can indeed solve the problem: increasing the size of decoding space generally leads to improving the BLEU score. We attribute this to the ability of the combined objective to measure both fluency and adequacy of translation candidates.
There is a significant gap between $k=10$ and $k=100$. However, keeping increasing $k$ does not result in significant improvements of translation accuracy but greatly decreases decoding efficiency.
Therefore, in the following experiments we set the max value of $k$ to $100$, and use normalized likelihood for $k=100$ if we don't use reconstruction in testing.


\subsection{Main Results}
\label{sec-results}

Table~\ref{table-translation-results} shows the translation performances on test sets measured in BLEU score. 
\textsc{RNNSearch} significantly outperforms Moses by 2.2 BLEU points on average, indicating that it is a strong baseline NMT system. 
This is mainly due to the introduction of two advanced techniques. Increasing beam size leads to decreasing translation performances on test sets, which is consistent with the result on the validation set.
We compare our methods with ``\textsc{RNNSearch} (Beam=10)'' in the following analysis, since it yields the best performance in the baseline systems.

First, the introduction of reconstruction significantly improves the performance over baseline by 1.1 BLEU points with beam size $k=10$. Most importantly, we obtain a further improvement of 1.2 BLEU points when expanding the decoding space. 
Second, our approach also consistently improves the quality (in terms of {\em Oracle} score, see the last column) of $k$-best translation candidates over the baseline system on various beam sizes. This confirms our claim that the combined objective contributes to parameter training for generating better translation candidates.

\subsection{Analysis}

We conduct extensive analyses to better understand our model in terms of efficiency of the added reconstruction, contribution of reconstruction from training and testing, alleviating typical translation problems, and building the ability of handling long sentences.

\paragraph{Speed}
Introducing reconstruction significantly slows down the training speed, while it slightly decreases the decoding speed.
For training, when running on a single GPU device Tesla K80, the speed of the baseline model is 960 target words per second, while the speed of the proposed model is 500 target words per second.
For decoding with beam=10, the speed of the baseline model is 2.28 seconds per sentence, while that of the proposed approach is  2.60 seconds per sentence.\footnote{For decoding with beam=100, the speeds are 22.97 and 25.29 seconds per sentence, respectively.} We attribute the effectiveness of decoding to the avoidance of beam search for reconstruction and the benefit of batch computation on GPU.

\begin{table}[h]
\setcounter{table}{4}
\centering
\renewcommand\arraystretch{1.1}
\begin{tabular}{c|c||c|c}
    \multicolumn{2}{c||}{\bf Rec. used in}	&	\multicolumn{2}{c}{\bf Beam}\\
    \hline
    {\em Training}	&	{\em Testing}	&	{$10$}	&	{$100$}\\
    \hline
    \texttimes	&	\texttimes	&	30.65		&	27.97\\
    \checkmark	&	\texttimes	&	31.17		&	31.51\\
    \checkmark	&	\checkmark	&	31.73		&	32.94\\
\end{tabular}
\caption{Contributions of reconstruction from parameter {\em training} and reranking of candidates in {\em testing}.}
\label{table-contribution}
\vspace{-15pt}
\end{table}

\paragraph{Contribution Analysis}
The contribution of reconstruction is of two-fold: (1) enabling parameter training for generating better translation candidates, and (2) enabling better reranking of generated candidates in testing. Table~\ref{table-contribution} lists the improvements from the two contribution sources.
When applied only in training, reconstruction improves translation performance by generating fluent and adequate translation candidates.
On top of that, reconstruction-based reranking further improves the performance.
The improvements are more significant when decoding spaces increase.

\begin{table}[h]
\centering
\renewcommand\arraystretch{1.1}
\begin{tabular}{l|c|c}
    {\bf Model}		&	{\bf Under-Tran.}	&	{\bf Over-Tran.}\\
    \hline
    \textsc{RNNSearch}			&	18.2\%	&	3.9\%\\
    \textsc{RNNSearch}+Rec.		&	16.2\%	&	2.4\%\\
\end{tabular}
\caption{Subjective evaluation of translation problems. Numbers denote percentages of source words.}
\label{table-subjective-evaluation}
\vspace{-15pt}
\end{table}

\paragraph{Problem Analysis}
We then conduct a subjective evaluation to investigate the benefit of incorporating reconstruction on the randomly selected 200 sentences. Table~\ref{table-subjective-evaluation} shows the results of subjective evaluation on translation.
\textsc{RNNSearch} suffers from serious under-translation and over-translation problems, which is consistent with the finding in other work~\cite{Tu:2016:ACL}.
Incorporating reconstruction significantly alleviates these problems, and reduces 11.0\% and 38.5\% of under-translation and over-translation errors respectively. The main reason is that both under-translation and over-translation lead to lower reconstruction scores, and thus are penalized by the reconstruction objective. As a result, the corresponding candidate is less likely to be selected as the final translation.

\begin{figure}[h]
\begin{center}
            \includegraphics[width=0.4\textwidth]{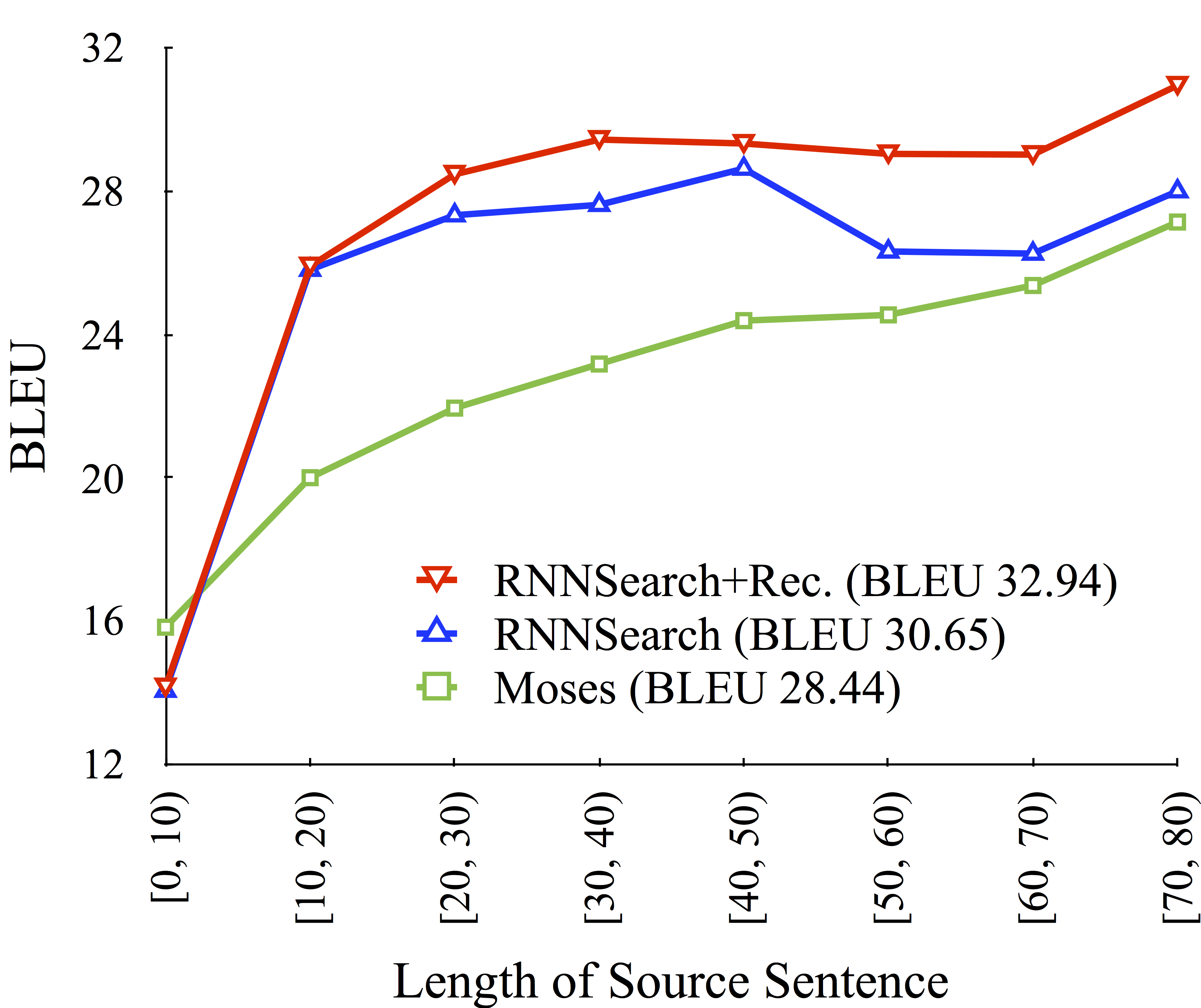}
      \caption{Performance of the generated translations with respect to the lengths of the input sentences on the test sets.} 
    \label{figure-sentence-len}
  \end{center}
  \vspace{-15pt}
\end{figure}

\paragraph{Length Analysis}
Following Bahdanau et al.~\shortcite{Bahdanau:2015:ICLR}, we group sentences of similar lengths together and compute the BLEU score for each group, as shown in Figure~\ref{figure-sentence-len}. Clearly the proposed approach outperforms all the other systems in all length segments.  
Specifically, \textsc{RNNSearch} outperforms Moses on all sentence segments, while its performance degrades faster than its competitors, which is consistent with the finding in~\cite{Bentivogli:2016:EMNLP}.
This is mainly due to that \textsc{RNNSearch} seriously suffers from inadequate translations on long sentences~\cite{Tu:2016:ACL}.
Our model explicitly encourages the decoder to incorporate source information as much as possible, and thus the improvements are more significant on long sentences.

\subsection{Comparison with Previous Work}

\begin{table}[t]
\centering
\renewcommand\arraystretch{1.1}
\begin{tabular}{l|c|c}
    {\bf Model}						&	{\bf Test}	&	{\bf $\bigtriangleup$}\\
    \hline
    \textsc{RNNSearch}				&	30.65		&	\\
    \hline
    \textsc{RNNSearch}+Cov.			&	31.89		&	\\
    \textsc{RNNSearch}+Cov.+{\em Rec.}		&	33.44	&	+1.6\\
    \hline
    \textsc{RNNSearch}+Ctx.			&	32.05	&	\\
    \textsc{RNNSearch}+Ctx.+{\em Rec.}		&	33.51	&	+1.5\\
    \hline
    \textsc{RNNSearch}+Cov.+Ctx.		&	33.12	&	\\
    \textsc{RNNSearch}+Cov.+Ctx.+{\em Rec.}	&	34.09		&	+1.0\\
\end{tabular}
\caption{Comparison with previous work on enhancing adequacy of NMT. ``Cov.'' denotes coverage mechanism to keep track of the attention history~\cite{Tu:2016:ACL}, and ``Ctx.'' denotes context gate to dynamically control the ratios at which source and target contexts contribute to the generation of target words~\cite{Tu:2016:arXiv}. 
}
\label{table-coverage}
\end{table}

We re-implement the methods of~\citeauthor{Tu:2016:ACL}~\shortcite{Tu:2016:ACL,Tu:2016:arXiv} on top of \textsc{RNNSearch}. For the coverage mechanism~\cite{Tu:2016:ACL}, we use the neural network based coverage, and the coverage dimension is 100. For the context gates~\cite{Tu:2016:arXiv}, we apply them on both source and target sides.
Table~\ref{table-coverage} lists the comparison results. Coverage mechanism and context gates significantly improve translation performance individually, and combining them achieves a further improvement. This is consistent with the results in~\cite{Tu:2016:ACL,Tu:2016:arXiv}.
Our model consistently improves the translation performance when further combined with the models.

\section{Related Work}

Our work is inspired by research on improving NMT by:
\paragraph{Enhancing Translation Adequacy}
Recently, several work shows that NMT favors fluent but inadequate translations~\cite{Tu:2016:ACL,Tu:2016:arXiv}.
While all the work is towards enhancing adequacy of NMT, 
our approach is complimentary: the above work is still under the standard encoder-decoder framework, while we propose a novel encoder-decoder-reconstructor framework.
Experiments show that combining those models together can further improve the translation performance.
\paragraph{Improving Beam Search}
Standard NMT models exploit a simple beam search algorithm to generate the translation word by word.
Several researchers rescore word candidates with additional features, such as language model probability~\cite{Gulcehre:2015:arXiv} and SMT features~\cite{He:2016:AAAI,Stahlberg:2016:arXiv}.
In contrast,~\citeauthor{Li:2016:NAACL}~\shortcite{Li:2016:NAACL} rescore translation candidates on sentence-level with the mutual information between source and target sides.
In the above work, NMT is treated as a black-box and its weighted outputs are combined with other features only in testing.
In this work, we move forward further by incorporating reconstruction score into the objective of training, which leads to creation of better translation candidates.
\paragraph{Capturing Bidirectional Dependency}
Standard NMT models only capture the unidirectional dependency from source to target with the likelihood objective.
It has been shown that combination of two directional models outperforms each model alone~\cite{Liang:2006:NAACL,Cheng:2016:IJCAI,Cheng:2016:ACL}.
Among them,~\citeauthor{Cheng:2016:ACL}~\shortcite{Cheng:2016:ACL} reconstruct the monolingual corpora with two separate source-to-target and target-to-source NMT models.
Closely related to~\citeauthor{Cheng:2016:ACL}~\shortcite{Cheng:2016:ACL}, our approach aims at enhancing adequacy of unidirectional (\ie source-to-target) NMT via an auxiliary target-to-source objective on parallel corpora, while theirs focuses on learning bidirectional NMT models via auto-encoders on monolingual corpora. Therefore, we use the decoder states as the input of the reconstructor, to encourage the target representation to contain the complete source information to reconstruct back to the source sentence.

\section{Conclusion}
We propose a novel encoder-decoder-reconstructor framework for NMT, in which the newly added reconstructor introduces an auxiliary score to measure the adequacy of translation candidates. 
The advantage of the proposed approach is of two-fold. First, it improves parameter training for producing better translation candidates. Second, it consistently improves translation performance when the decoding space increases, while conventional NMT fails to do so. Experimental results show that the two advantages can indeed help our approach to consistently improve translation performance.

There is still a significant gap between de facto translation and oracle of $k$-best translation candidates, especially when the decoding space increases. We plan to narrow the gap with rich features, which can better measure the quality of translation candidates.
It is also necessary to validate the effectiveness of our approach on more language pairs and other NMT architectures.

\section*{Acknowledgement}

This work is supported by China National 973 project 2014CB340301.
Yang Liu is supported by the National Natural Science Foundation of China (No. 61522204) and the 863 Program (2015AA015407).

\bibliography{all}
\bibliographystyle{aaai}

\end{document}